\documentclass[a4paper, 10pt, conference]{ieeeconf}
\usepackage[a4paper, total={7in, 9.5in}, bottom=0.75in, left=0.75in]{geometry}
\IEEEoverridecommandlockouts

\usepackage{amsmath,amssymb,amsfonts}
\usepackage{dsfont}
\usepackage[linesnumbered,ruled,vlined]{algorithm2e}
\usepackage{graphicx}
\usepackage{subcaption}
\usepackage{textcomp}
\usepackage{xcolor}
\usepackage{soul}
\usepackage{bm}
\usepackage{siunitx}
\usepackage{esvect}
\usepackage{xfrac}
\usepackage{hyperref}
\hypersetup{
    colorlinks=true,
    citecolor=black,
    linkcolor=black,
    urlcolor=blue
}

\newcommand{\vect}[1]{\boldsymbol{\mathbf{#1}}}

\newcommand{\onevec}[1]{\vect{\mathds{1}_{#1}}}

\newcommand{\force}{f}

\let\oldphi\phi \let\phi\varphi \let\varphi\oldphi

\title{\LARGE \bf
Trajectory Optimization for Thermally-Actuated Soft Planar Robot Limbs
}
\author{Anthony Wertz$^{1*}$, Andrew P. Sabelhaus$^{2*}$, Carmel Majidi$^{1,3}$% <- stops a space
\thanks{This work was in part supported by the Office of Naval Research under Grant No. N000141712063 (PM: Dr. Tom McKenna), the National Oceanographic Partnership Program (NOPP) under Grant No. N000141812843 (PM: Dr. Reginald Beach), and an Intelligence Community Postdoctoral Research Fellowship through the Oak Ridge Institute for Science and Education.}
\thanks{$^1$A. Wertz and C. Majidi are with the Robotics Institute, Carnegie Mellon University, Pittsburgh PA, USA. {\tt\small awertz, cmajidi@andrew.cmu.edu} }
\thanks{$^2$A.P. Sabelhaus was with the Department of Mechanical Engineering, Carnegie Mellon University, Pittsburgh PA, USA. He is now with the Department of Mechanical Engineering, Boston University, Boston MA 02215 {\tt\small asabelha@bu.edu}}
\thanks{$^3$C. Majidi is also with the Department of Mechanical Engineering, Carnegie Mellon University, Pittsburgh PA, USA.}
\thanks{$^*$A. Wertz and A.P. Sabelhaus contributed equally to this work.}
}

\begin{document}
\bstctlcite{BSTcontrol}

\maketitle
\thispagestyle{empty}
\pagestyle{empty}

\begin{abstract}
Practical use of robotic manipulators made from soft materials requires generating and executing complex motions.
We present the first approach for generating trajectories of a thermally-actuated soft robotic manipulator.
Based on simplified approximations of the soft arm and its antagonistic shape-memory alloy actuator coils, we justify a dynamics model of a discretized rigid manipulator with joint torques proportional to wire temperature.
Then, we propose a method to calibrate this model from experimental data and demonstrate that the simulation aligns well with a hardware test.
Finally, we use a direct collocation optimization with the robot's nonlinear dynamics to generate feasible state-input trajectories from a desired reference.
Three experiments validate our approach for a single-segment robot in hardware: first using a hand-derived reference trajectory, then with two teach-and-repeat tests.
The results show promise for both open-loop motion generation as well as for future applications with feedback.
\end{abstract}

%%%%%%%%%%%%%%%%%%%%%%%%%%%%%%%%%%%%%%%%%%%%%%%%%%%
%%%%%%%%%%%%%%%%%%%%%%%%%%%%%%%%%%%%%%%%%%%%%%%%%%%
%%%%%%%%%%%%%%%%%%%%%%%%%%%%%%%%%%%%%%%%%%%%%%%%%%%
\section{Introduction}

Soft robots may outperform their rigid counterparts in tasks requiring biomimetic deformability and safe, robust environmental interaction \cite{laschi_soft_2016,majidi_soft_2014}.
However, practical use of soft robots requires performing similarly-advanced motions as rigid robots.
Many soft robots struggle to match more complicated trajectories due to limitations in actuation, design~\cite{Rodrigue2017}, and modeling \cite{bruder_nonlinear_2019} for high-degree-of-freedom state spaces.
Generating feasible motions, and corresponding inputs, requires tractable models that accurately reproduce hardware behavior.
This is particularly challenging for soft robots actuated with thermally-responsive materials like shape-memory alloy (SMA) becoming increasingly popular within the field \cite{huang2020shape,el2020soft}.
Though shape-memory actuators require minimal added hardware while providing high work density, unlike cable-driven and pneumatic systems, they are especially difficult to model and simulate for robotic applications~\cite{ge_preisach-model-based_2021}.

\begin{figure*}[tph]
    \centering
    \includegraphics[width=1\textwidth]{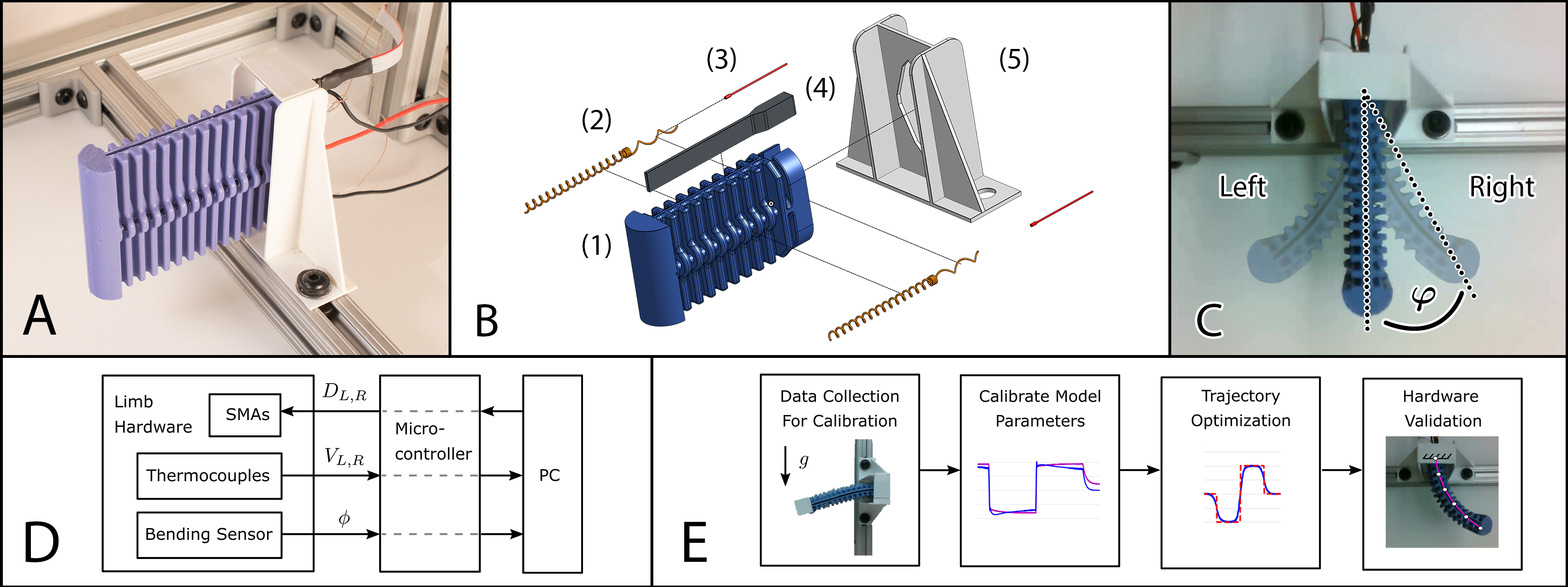}
    \caption{
    The limb (A) consists of cast bulk silicone (B1) actuated by two antagonistic SMA coils (B2).
    Thermocouples (B3) and a bend sensor (B4) are included.
    A bracket (B5) holds the limb horizontally to remove gravitational loading on the bending axis. 
    Bending (C) is achieved using PWM signals from a microcontroller (D) to actuate the SMA coils through Joule heating.
    (E) Our procedure collects hardware data, calibrates the model, optimizes trajectories, then validates on hardware.}
    \vspace{-0.4cm}
    \label{fig:overview}
\end{figure*}

We propose a modeling and trajectory generation framework for soft limbs with antagonistic thermal actuators (Fig. \ref{fig:overview}).
From prior work and first-principle approximations we develop a three-part dynamics model based on a rigid manipulator, a map from actuator temperature to joint torque, and Joule heating.
We construct a single segment of a soft limb with two SMA actuators, motivated by \cite{patterson_untethered_2020} with the addition of temperature and displacement sensing, and calibrate the model using our hardware.
We develop and solve an optimization problem to generate feasible state and control trajectories.
Open-loop hardware tests demonstrate that our approach can be used to re-create ``teach and repeat'' motions of the limb.

We focus on open-loop operation, as feedback control has been extensively studied for soft robots with thermal \cite{yang_design_2019,luong_long_2021} and controllable-force actuators \cite{marchese_dynamics_2016,della2020model}.
Trajectory optimization offers two distinct benefits in comparison to feedback without pre-planned trajectories~\cite{kelly2017introduction}:
First, dynamic feasibility is verified \textit{a~priori}, a challenging requirement for many state-feedback techniques (e.g., model-predictive control~\cite{sabelhaus_model-predictive_2021}) in soft and flexible robots.
Second, objectives of interest, like minimum energy expenditure or minimum time, can be incorporated directly through costs or constraints.
Since thermal actuators have significant energy requirements~\cite{Rodrigue2017}, these are often opposing goals.

This article contributes a trajectory optimization technique for a thermally-actuated soft planar robot limb, comprising:

\begin{enumerate}
    \item An approximated model of the manipulator and actuator dynamics with low computational cost,
    \item An optimization routine that uses the model for generating feasible trajectories, and
    \item A validation of the approach, faithfully re-creating three motions of the limb in hardware.
\end{enumerate}

This article applies our approach to a single-segment robot with two actuators. 
Single-segment planar soft robots are found in a variety of settings, such as multi-fingered hands with a rigid base \cite{turco_grasp_2021}, and are routinely used as benchmarks \cite{tang_probabilistic_2020}.
Our approach makes motion planning possible when thermal actuators are used in these applications.
This is the first demonstration of a re-usable computational trajectory generation approach for any thermally-actuated soft robot.

%%%%%%%%%%%%%%%%%%%%%%%%%%%%%%%%%%%%%%%%%%%%%%%%%%%
%%%%%%%%%%%%%%%%%%%%%%%%%%%%%%%%%%%%%%%%%%%%%%%%%%%
%%%%%%%%%%%%%%%%%%%%%%%%%%%%%%%%%%%%%%%%%%%%%%%%%%%
\section{Related work}

We consider soft limbs actuated with SMA coils that, due to their size and work density advantages \cite{haines2014artificial,majidi2019soft}, are promising for compact, untethered robots \cite{patterson_untethered_2020}.
However, other soft actuation mechanisms (such as pneumatics and cables) can directly control applied force \cite{marchese_dynamics_2016,graule_somo_2021}, whereas thermal actuation only occurs indirectly \cite{sabelhaus2021gaussian}.
Since traditional robotics models do not capture thermal actuator dynamics, little progress has been made toward dynamic trajectory generation for these mechanisms.
Prior work includes \(A*\) to optimize SMA arrays \cite{mollaei2012optimal} and evolutionary algorithms for rigid SMA-actuated robots \cite{katoch2015trajectory}.
Neither test their results on hardware, nor address feasibility.
There have been attempts at open-loop SMA task-space operation \cite{bena2021smarti}, though not for soft manipulators.
To our knowledge, no prior work has computationally generated feasible state-input trajectories of a soft robot limb with thermal actuators.

Many models of soft-bodied robot arm kinematics and dynamics are available \cite{sadati2017mechanics} with trade-offs between complexity, computational requirements, and physical accuracy.
More physically accurate models include the discrete elastic rod (DER) method \cite{goldberg2019planar,huang2020dynamic,huang2021numerical}.
More computationally-tractable models include the constant-curvature framework \cite{webster2010design,della2020model}.
We use one of the simplest possible representations: the rigid manipulator, motivated by approximations of the above alongside promising contemporary results \cite{graule_somo_2021}.

Dynamics for thermal actuators are often based on first principles and constitutive models \cite{cheng2017modeling,haines2014artificial}.
However, these often require measurements of stress and strain \cite{ge_preisach-model-based_2021}, which adds to computational complexity and presents a challenge for robot design.
This article investigates if temperature alone can approximate a more complicated stress/strain actuator response, since temperature is more readily modeled with
Joule heating and convective cooling \cite{bhargaw2013thermo,sabelhaus2021gaussian}.
Our simple temperature-to-stress model makes computational trajectory generation possible for this class of robots.

%%%%%%%%%%%%%%%%%%%%%%%%%%%%%%%%%%%%%%%%%%%%%%%%%%%
%%%%%%%%%%%%%%%%%%%%%%%%%%%%%%%%%%%%%%%%%%%%%%%%%%%
%%%%%%%%%%%%%%%%%%%%%%%%%%%%%%%%%%%%%%%%%%%%%%%%%%%
\section{Hardware platform}

Our soft limb (Fig. \ref{fig:overview}A, B) is derived from prior work in a legged robot \cite{patterson_untethered_2020}, with the intent to eventually be employed in that setting.
The limb body was cast from bulk silicone elastomer (Smooth-Sil 945, Smooth-On) and embedded with nickel-titanium alloy SMA actuator coils (Flexinol, Dynalloy) and a capacitive bend sensor (single axis, Bend Labs).
Thermocouples for measuring SMA wire temperature were bonded to the bracketed side of the actuators with thermally conductive, electrically insulating epoxy (MG 8329TCF).
The limb was mounted with a 3D-printed bracket oriented with the bending axis parallel to gravity (Fig. \ref{fig:overview}) so gravitational loading can be ignored.

Sensing and control were performed with an offboard microcontroller (nRF52-DK, Nordic Semiconductor).
The thermocouples (attached to an amplifier, MAX31855) and the bending sensor both communicated digitally with the microcontroller (Fig. \ref{fig:overview}D).
Current through the SMA actuators was controlled using pulse-width modulation (PWM) to N-channel power MOSFETs connected to a 7V power supply. 

%%%%%%%%%%%%%%%%%%%%%%%%%%%%%%%%%%%%%%%%%%%%%%%%%%%
%%%%%%%%%%%%%%%%%%%%%%%%%%%%%%%%%%%%%%%%%%%%%%%%%%%
%%%%%%%%%%%%%%%%%%%%%%%%%%%%%%%%%%%%%%%%%%%%%%%%%%%
\section{Dynamics modeling}

The approach in this article makes approximations to each of the three relevant dynamics phenomena of our robot limb: the manipulator body, the discretized joint torques, and the actuator temperature.
These are combined into a final set of equations of motion, taking a pulse-width modulation (PWM) voltage on the SMA wires as input and predicting the manipulator's bending angle as output.

%%%%%%%%%%%%%%%%%%%%%%%%%%%%%%%%%%%%%%%%%%%%%%%%%%%
\subsection{Rigid manipulator model}

\begin{figure} 
    \centering
    \includegraphics[width=0.9\linewidth]{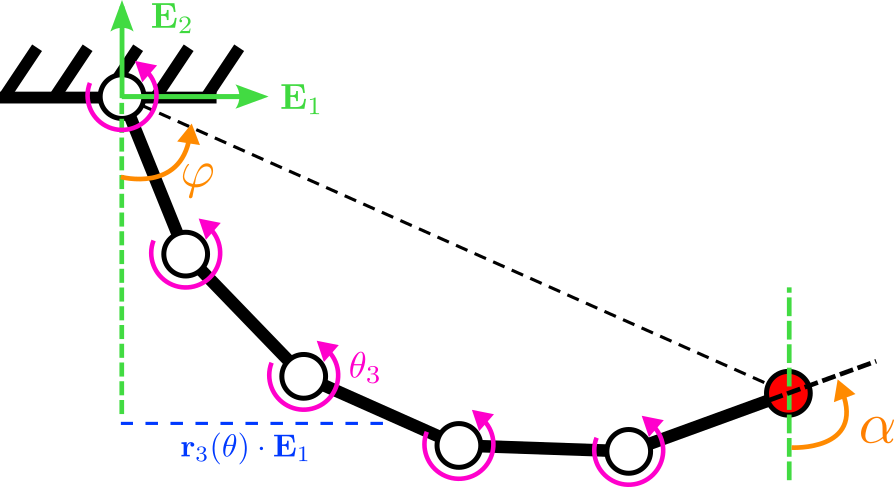}
    \caption{Fixed-base manipulator model with uniformly spaced revolute joints rotating normal to the plane. Orthogonal basis vectors \(\vect{E}_1\) and \(\vect{E}_2\) define the inertial frame centered on the first joint. The bend angle \(\phi\) is estimated from the sensor's tip tangent angle $\alpha$ (eqn. \ref{eqn:phi}) for model calibration (Sec. \ref{sec:cal}) along with link center of mass offsets (e.g.,  \(\vect{r_3}(\vect{\theta})\cdot\vect{E_1}\)).}
    \label{fig:model:manip}
\end{figure}

This article employs a simplified model of a discretized rigid manipulator for the robot's body.
Two observations motivate this approach.
First, recent work has demonstrated relatively accurate simulations of soft fluid-driven limbs as discretized manipulators \cite{graule_somo_2021}.
Second, the discretized rigid manipulator arises from the discrete elastic rod (DER) model, previously shown to accurately model SMA-driven soft robots \cite{goldberg2019planar,huang2020dynamic}, under certain approximating assumptions.
In particular, our robot does not experience significant centerline extension. 
The DER with no stretching is dynamically equivalent to a serial-chain rigid manipulator with nonlinear springs at each discretized joint.
Then, modeling the bending forces with \textit{linear} springs instead, the dynamics become the flexible manipulator model (Fig. \ref{fig:model:manip}) of

\begin{align}\label{eqn:flex_manip}
    \vect{M}(\vect{\theta})\vect{\ddot\theta} + \vect{C}(\vect{\theta},\vect{\dot\theta})\vect{\dot\theta} + k(\vect{\theta} - \vect{\bar\theta}) + \sigma\vect{\dot\theta} = \vect{0},
\end{align}

\noindent with the conventional mass and Coriolis/centrifugal terms \(M\), \(C\) and spring constant $k$.
We include linear damping \(\sigma\) for energy dissipation in the soft material as in \cite{huang2020dynamic}.
The vector $\vect{\theta}$ are angles between the $n$ discretized segments: $\vect{\theta} = [\theta_1, \; \theta_2 \;, \hdots, \theta_n]$, hereafter referred to as joint angles.

The discretized manipulator model in eqn. (\ref{eqn:flex_manip}) actuates by changing the set-point angle $\vect{\bar\theta}$ of the torsional springs, as in the DER dynamics \cite{huang2020dynamic}.
Moreover, as in \cite{huang2020dynamic}, we assume that this change occurs as a function of temperature in our two SMAs, \(\vect{T}\) = $[T_l, T_r]$, where \(r\) indicates the right-side actuator and \(l\) the left-side actuator. 
Since $k$ distributes through our linear spring, we re-write the generalized force (generalized torque) due to actuation as $\vect{\force}(\vect{T}) = k \vect{\bar\theta}(\vect{T})$, to be specified and calibrated later, arriving at

\begin{align}
    & \vect{M}(\vect{\theta})\vect{\ddot\theta} + \vect{C}(\vect{\theta},\vect{\dot\theta})\vect{\dot\theta} + k\vect{\theta} + \sigma\vect{\dot\theta} = \vect{\force}(\vect{T}).  \label{eq:model:manip}
\end{align}

%%%%%%%%%%%%%%%%%%%%%%%%%%%%%%%%%%%%%%%%%%%%%%%%%%%
\subsection{SMA actuator model: temperature-force relationship}\label{sec:model:force}

Eqn. (\ref{eq:model:manip}) takes $\vect{f}(\vect{T})$ to be a static stateless mapping, i.e., neglects the internal constitutive properties of the SMA in favor of reduced computational complexity.
To justify this highly simplified relationship, we consider the constitutive model for a single SMA, and determine what approximations are implied.
The SMA strain-stress-temperature model is well known in the literature, for example, as in~\cite{cheng2017modeling} for one wire,

\begin{align}
    \tau - \tau_0 &= G(\xi)(\gamma - \gamma_0) + \frac{\Theta}{\sqrt{3}}(T - T_0) + \frac{\Omega(\xi)}{\sqrt{3}}(\xi - \xi_0)
\end{align}

\noindent with the wire's shear stress \(\tau\), shear modulus \(G\), shear strain \(\gamma\), coefficient of thermal expansion \(\Theta\), temperature \(T\), phase transformation coefficient \(\Omega\), and martensite fraction \(\xi\).
Quantities with \textit{naught} subscript, e.g., \(\tau_0\), are the values at ambient temperature.
In order, if the following approximations are assumed:

\begin{enumerate}
    \item Thermal expansion is negligible (as in \cite{cheng2017modeling}),
    \item Change in strain is small in comparison to change in stress, i.e., $G(\xi)(\gamma - \gamma_0) << (\tau - \tau_0)$ for our range of interest in $\xi$, so $G(\xi)(\gamma - \gamma_0) \approx 0$,
    \item Martensite fraction is proportional to wire temperature, \(\xi \propto T - T_0\),
\end{enumerate}

\noindent then \(\tau \propto T - T_0\).
Lastly, assuming that \(\tau\) acts uniformly across joints and that it induces the generalized torques in eqn. (\ref{eq:model:manip}), then \(\vect{f} \propto \vect{\tau}\).
We do not consider any center-line dependence as would be expected in a tendon-driven system~\cite{camarillo2008mechanics}; rather, this is accounted for via calibration of the torsional spring constant in the manipulator model.
Lumping each scaling factor into a parameter $\beta \in \mathbb{R}$, then \(\vect{\force} = \beta(T - T_0)\onevec{n}\), where \(\onevec{n}\) is a vector of ones of length \(n\).
Therefore, with two SMA actuators acting antagonistically,

\begin{align}\label{eq:model:forcetemp}
    \vect{\force}(\vect{T}) &= \beta_r(T_r - T_0)\onevec{n} - \beta_l(T_l - T_0)\onevec{n}.
\end{align}

\noindent These assumptions are simplistic, but our hardware validation testing suggests they are a useful approximation.

%%%%%%%%%%%%%%%%%%%%%%%%%%%%%%%%%%%%%%%%%%%%%%%%%%%
\subsection{SMA actuator model: temperature dynamics}\label{sec:model:smatemp}

Finally, we design a relationship between our control inputs, the PWM duty cycles $\vect{u} = [D_l, D_r] \in [0, 1]^{2}$, and the SMA wire temperatures.
As with stress vs. temperature, we consider each wire individually.
It has been shown in the literature \cite{mollaei2012optimal,katoch2015trajectory,bhargaw2013thermo} that SMA wire temperature can be approximately modeled by Joule heating in the form

\begin{align}
    \dot T &= -\frac{h_cA_c}{C_v}(T - T_0) + \frac{1}{C_v} P
\end{align}

\noindent with specific heat capacity \(C_v\), ambient heat convection coefficient \(h_c\), surface area \(A_c\), ambient temperature \(T_0\), and input electrical power $P$. 
For current-controlled SMAs, $P=\rho J^2$, where $\rho$ is resistance and $J$ is current density.
For our PWM input, we assume that the duty cycle \(D\) modulates the fraction of time current is conducting through the SMA and that current is constant when flowing, so $P=\rho J^2 D$.

Our embedded thermocouple is bonded to the SMA wire using a small amount of thermally conductive, electrically insulating epoxy; this adds thermal mass.
We therefore model the measured temperature \(V\) with an additional linear time delay.
As a result, both $\dot T$ and $\dot V$ are linear systems, of the form

\begin{align}
    \dot T &= a_1(T - T_0) + a_2D\,, \label{eq:Tdot} \\
    \dot V &= a_3(V - T)\,, \label{eq:Vdot}
\end{align}

\begin{figure*}[th]
    \centering
    \includegraphics[width=0.9\textwidth]{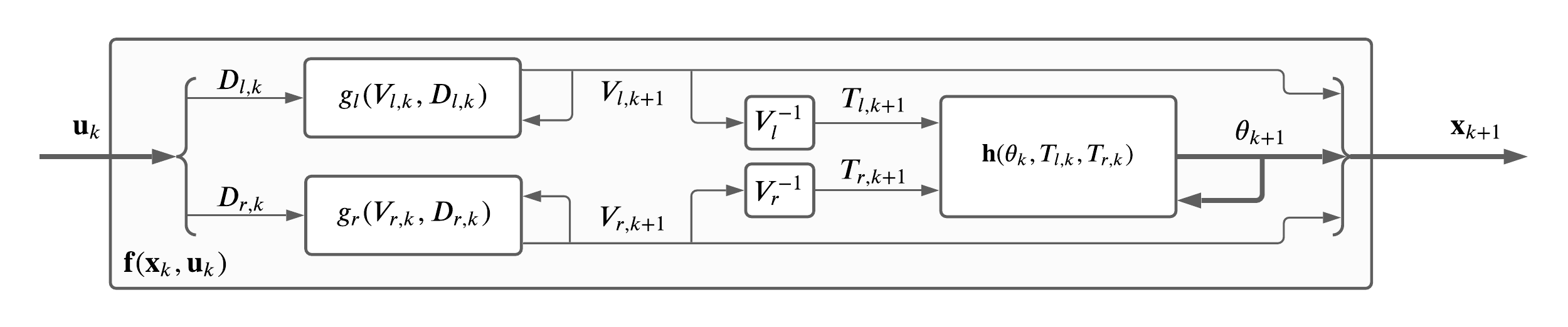}
    \caption{The combined dynamics \(f\) is composed of two thermal actuator blocks \(g_l,g_r\), eqns. (\ref{eq:Tdot})-(\ref{eq:Vdot}), and the serial manipulator dynamics \(h\), eqn. (\ref{eq:model:manip}). Wire temperature is obtained from measured temperature (\(V_j^{-1}\)) via $T_j = V_j - \dot V_j/a_3$.}
    \label{fig:model:block}
    \vspace{-0.1cm}
\end{figure*}

\noindent where \(a_1=-h_cA_c/C_v\) and \(a_2=\rho J^2/C_v\).
This linear model is equivalent to that used in \cite{luong_long_2021}.
The measurement temperature \(V\) is the state tracked in the simulated system dynamics, but \(T\) is readily computed by rearranging the terms in eqn. \ref{eq:Vdot}.
The full model dynamics are depicted in Fig. \ref{fig:model:block}.

%%%%%%%%%%%%%%%%%%%%%%%%%%%%%%%%%%%%%%%%%%%%%%%%%%%
%%%%%%%%%%%%%%%%%%%%%%%%%%%%%%%%%%%%%%%%%%%%%%%%%%%
%%%%%%%%%%%%%%%%%%%%%%%%%%%%%%%%%%%%%%%%%%%%%%%%%%%
\section{Model Calibration} \label{sec:cal}

The dynamics model of eqns. (\ref{eq:model:manip}, \ref{eq:model:forcetemp}, \ref{eq:Tdot}, \ref{eq:Vdot}) requires calibration from data.
Physical parameters (mass, mass moments of inertia, geometry) were measured using laboratory scales and our CAD model.
For the remainder, we make the simplifying assumption of \textit{constant curvature}~\cite{webster2010design} to map the angular displacement output of our bending sensor, \(\alpha\), to the bending angle of the limb, \(\phi\) (Fig. \ref{fig:model:manip}), which is 

\begin{equation}
    \phi=\alpha/2. \label{eqn:phi}
\end{equation}

\noindent We first calibrate the two passive dynamics parameters, spring constant $k$ and damping constant $\sigma$, then use actuated data to calibrate the generalized force and temperature models.

%%%%%%%%%%%%%%%%%%%%%%%%%%%%%%%%%%%%%%%%%%%%%%%%%%%
\subsection{Linear torsional spring constant}

To calibrate the spring constants \(k\), we reoriented the limb with the bending axis parallel to the ground and measured the deflection in static equilibrium under gravitational loading.
Here, with $\vect{\dot\theta}^{eq}=\vect{0}$ and $\vect{f}(\vect{T}^{eq})=\vect{0}$, the manipulator dynamics (eqn. \ref{eq:model:manip}) simplify to \(k\vect\theta^{eq} + \vect{\force_g}(\vect\theta^{eq}) = \vect{0}\), picking up a new term due to gravity.
We use an exponential map for the kinematics of the link centers of mass \(\vect{r}_i(\vect\theta^{eq})\).
Gravitational potential energy is then \(U_g(\vect{\theta}^{eq}) = mg\sum_{i=1}^{n}\vect{r_i}(\vect{\theta}^{eq})\cdot \vect{E_1}\), and so we computed \(\vect{\force_g}(\vect{\theta}^{eq}) = -\nabla_{\vect{\theta}} U_g(\vect{\theta}^{eq})\).
A least-squares fit then gives \(k = -\vect{\theta}^{eq}\setminus\vect{\force_g}(\vect\theta^{eq})\) from hardware data of $\vect{\theta}^{eq}$.

However, our bend sensor only provides a scalar measurement of $\phi^{eq}$, not the joint angles $\vect{\theta}^{eq}$, and the constant curvature assumption needed for eqn. (\ref{eqn:phi}) is only a rough approximation for our manipulator under gravity.
To obtain $\theta^{eq}_i$ from $\phi^{eq}$, we treat our manipulator as a discretized version of an Euler-Bernoulli beam under gravitational loading. 
The comparable loading condition is a moment applied at each joint, arising from gravitational generalized force at links further along the cantilever. 
A static equilibrium calculation, augmented with a correction factor \(\lambda \in \mathbb{R}^+\) to partially account for eqn. (\ref{eqn:phi})'s approximation, gives

\begin{equation}\label{eqn:beamexponential}
    \theta^{eq}_i = \lambda \phi^{eq} \frac{(N-i+1)^2}{\sum_{j=1}^{N}(N-j)^2} = \lambda \phi^{eq} b_i
\end{equation}

\noindent Given an observation $\{\phi^{eq}, \vect{\theta}^{eq}\}$ from a simulation of eqn. (\ref{eq:model:manip}) as \(k\vect\theta^{eq} + \vect{\force_g}(\vect\theta^{eq}) = \vect{0}\), an estimate for the correction factor is $\lambda^* = \theta_i^{eq} / (\phi^{eq} b_i)$.
We iterated between calculating $k$ then re-estimating $\lambda^*$ via simulation, starting from $\lambda^*=1$ until both converged (at $\lambda^* \approx 1.22$).

%%%%%%%%%%%%%%%%%%%%%%%%%%%%%%%%%%%%%%%%%%%%%%%%%%%
\subsection{Damping constant}\label{sec:cal:damping}

To calibrate the damping constant $\sigma$, we displaced the tip of the limb to \ang{45}, released, and collected hardware data $\phi^d_{1 \hdots t}$ during passive oscillation until the limb came to rest.
With no actuation or external loading, eqn. (\ref{eq:model:manip}) simplifies to

\begin{align}
    \vect{M}(\vect{\theta})\vect{\ddot\theta} + \vect{C}(\vect{\theta},\vect{\dot\theta})\vect{\dot\theta} + k\vect{\theta} + \sigma\vect{\dot\theta} = \vect{0}\,. \label{eq:sigma:eom}
\end{align}

We use a nested, two-step optimization (Alg. \ref{alg:sigmaopt}) to estimate $\sigma$ without needing to map $\phi \rightarrow \vect{\theta}$.
First, we note that we can optimize the parameters of a scalar damped sinusoid to fit a bend angle trajectory $\phi_{1 \hdots t}$ as $\sigma^* = \texttt{fitDS}(\phi_{1 \hdots t})$.
However, we observed that a na\"ive use of $\sigma=\sigma^*_{d} =\texttt{fitDS}(\phi^d_{1 \hdots t})$ in simulations of eqn. (\ref{eq:sigma:eom}) overdamps the response.
The outer loop of our optimization, \texttt{fitData}, therefore minimizes the difference between the damping constant estimate from hardware, $\sigma^*_d$, and the damping constant estimate from manipulator simulations $\sigma^*_{mdl} = \texttt{fitDS}(\phi^{mdl}_{1 \hdots t} | \sigma_{mdl})$.
Our implementation of \texttt{fitData} obtains $\phi_{1 \hdots t}^{mdl}$ by rolling out eqn. (\ref{eq:sigma:eom}) given a $\sigma_{mdl}$.
Therefore, $\sigma=\sigma^*_{mdl}$ is the manipulator (simulation) damping constant that best re-creates the magnitude of damping observed from a hardware fit.
All future simulations of eqn. (\ref{eq:model:manip}) used $\sigma^*_{mdl}$.

%%%%%%%%%%% ALGORITHM:
\begin{algorithm}[h]
    \caption{
    Nested optimization to find $\sigma$ by matching the damping observed in hardware data $\phi_{1 \hdots t}^d$.}\label{alg:sigmaopt}
    \SetKwInput{KwInput}{Input}                % Set the Input
    \SetKwInput{KwOutput}{Output}              % set the Output
    \DontPrintSemicolon
    
    % Set Function Names
    \SetKwFunction{FData}{fitData}
    \SetKwFunction{FDampedSin}{fitDS}
 
    % Write Function with word ``Function''
    \SetKwProg{Fn}{Procedure}{:}{}
    \Fn{\FDampedSin{\(\phi_{1 \hdots t}\)} \(\rightarrow \sigma^*\)}{
        \(\zeta^*, \omega^*_n \leftarrow \arg\min ||\phi_{1 \hdots t} - Ae^{-\zeta\omega_nt}\sin(\omega_n\sqrt{1-\zeta^2}t + \varphi) + b||^2_2\)\;
        \(\sigma^* \leftarrow \zeta^*\omega^*_n\)\;
    }
    
    \SetKwProg{Fn}{Procedure}{:}{\KwRet}
    \Fn{\FData{\(\phi^d_{1 \hdots t}\)} \(\rightarrow \sigma^*_{mdl}\)}{
        \(\sigma^*_{d} \leftarrow \texttt{fitDS}(\phi^d_{1 \hdots t})\)\;
        \(\sigma^*_{mdl} \leftarrow \arg\min ||\sigma^*_{d} - \texttt{fitDS}(\phi^{mdl}_{1 \hdots t} | \sigma_{mdl})||^2_2\)\;
    }
\end{algorithm}

%%%%%%%%%%%%%%%%%%%%%%%%%%%%%%%%%%%%%%%%%%%%%%%%%%%
\subsection{Thermal actuator constants}\label{sec:thermal_calibration}

Thermal actuator constants were calculated with the limb reoriented back (no gravity).
Here, the constant curvature assumption holds, and we can calculate $\theta_i$ from $\phi$ directly.
With some trigonometry, and assuming identical joint angles,

\begin{align}\label{eqn:beamcc}
    \theta_i &= 2\phi/(n+1), \quad \quad \forall i = 1\hdots n.
\end{align}

To calibrate the constants for heat transfer, we collected data at static equilibria where our actuators were heated.
Theoretically, if we operate our system at $\vect{f}(\vect{T}^{eq}) \neq \vect{0}$, $\dot \phi=0$, and $\vect{\dot V}=\vect{0}$, three conditions arise.
Assuming we have held equilibrium for a sufficient amount of time, the wire temperature is equal to the measured temperature, $\vect{V}^{eq}=\vect{T}^{eq}$.
Second, the dynamics in eqns. (\ref{eq:model:manip}), (\ref{eq:model:forcetemp}) reduce to 

\begin{equation}\label{eqn:heatedeq}
    k \vect{\theta}^{eq} = \frac{2k\phi^{eq}}{(n+1)}\onevec{n} = \left[ \beta_r(T^{eq}_r - T_0) - \beta_l(T^{eq}_l - T_0) \right]\onevec{n}
\end{equation}

\noindent and so a fixed temperature maps to one fixed robot pose.
Third, $\vect{V}^{eq} \neq0 \Rightarrow \phi^{eq} \neq 0$.
Together, these observations allow us to calibrate $\{a_{1,j}, a_{2,j}, a_{3,j}, \beta_j\}$ for both actuators $j$ without modeling dynamic motions.

We developed a simple PI feedback controller from $\phi$ to $\vect{D}$ to stabilize the limb around $\phi^{eq} \neq 0$.
Using this controller, we generated three calibration datasets of the form $\mathcal{C} = \{ \phi, \phi^{eq}, \vect{V}, \vect{D}\}_{1 \hdots t}$ by randomly selecting $\phi^{eq}$ values in a range and operating our controller for some time at each.
The first two datasets involved motion in which only one SMA was activated: $\mathcal{C}^r$ used $\phi^{eq} \in (0, \ang{45})$ where the controller applied $D_l = 0$, vice-versa for a set $\mathcal{C}^l$.
The third set $\mathcal{C}^m$ had both actuators activated.

\vspace{0.1cm}
\subsubsection{Heat transfer coefficients}

We independently fit the three parameters $\{a_{1j},a_{2j},a_{3j}\}$ in eqns. (\ref{eq:Tdot}) and (\ref{eq:Vdot}) for each SMA actuator $j$ using $\mathcal{C}^j$.
To obtain a time series for $T_{j, 1 \hdots t}$ from $\mathcal{C}^j$, we observe that eqn. (\ref{eqn:heatedeq}) implies equilibrium bending angle should be a linear scaling factor of temperature.
That is, rearranging eqn. (\ref{eqn:heatedeq}) becomes $T_j^{eq} = b \phi^{eq} + T_0$ for some $b \in \mathbb{R}$.
We examined $\mathcal{C}^j$ to find the most promising point $(\phi^{eq*}, V_j^{eq*})$ where $\dot V_j \approx 0$, then with some algebra, $b = (V^{eq*}-T_0)/\phi^{eq*}$.
Lastly, we used this static relationship as a rough approximation for dynamic wire temperature not at equilibrium, $T_{j, 1 \hdots t} = b \phi_{1 \hdots t} + T_0$.
With known trajectories for $T_j, V_j, D_j$, we used the \texttt{DiffEqParamEstim.jl} Julia package to fit the parameters, first using {\tt two\_stage\_method} (a two-stage collocation procedure) to find a rough estimate, then {\tt optimize} for refinement, both with an \(L_2\) residual loss.

\vspace{0.1cm}
\subsubsection{Actuator force coefficients}

To find $\beta_j$ using $\mathcal{C}^j$, we observe that eqn. (\ref{eqn:heatedeq}) also implies \(\beta_j = [k/(T^{eq}_{j} - T_0)]\theta^{eq}_i\).
We create a better estimate of $T_{j, 1 \hdots t}$ by simulating eqns. (\ref{eq:Tdot})-(\ref{eq:Vdot}) using our calibrated $a_{1 \hdots 3,j}$ on the data in $\mathcal{C}^j$.
Then, we picked out timepoints where $\dot \phi \approx 0$ in $\mathcal{C}^j$, and took an average $T^{eq}$ over a window around those points.
Since we know $k$, stacking all \(t\) observations in the vector \(\vect\Gamma^{eq}_j\) we have

\begin{align}
    \vect \Gamma^{eq}_j &= \begin{bmatrix}
        \left(k/(T^{eq}_{j1} - T_0)\right)\theta^{eq}_{i1} \\
        \vdots \\
        \left(k/(T^{eq}_{jt} - T_0)\right)\theta^{eq}_{it}
    \end{bmatrix}\,,
\end{align}

\noindent where we then computed a least-squares fit as \(\beta_j = \onevec{t}\setminus\vect{\Gamma}^{eq}_j\) to get force coefficients for each actuator $j$.

%%%%%%%%%%%%%%%%%%%%%%%%%%%%%%%%%%%%%%%%%%%%%%%%%%%
\subsection{Model Validation}
\begin{figure} 
    \centering
    \includegraphics[width=1\linewidth]{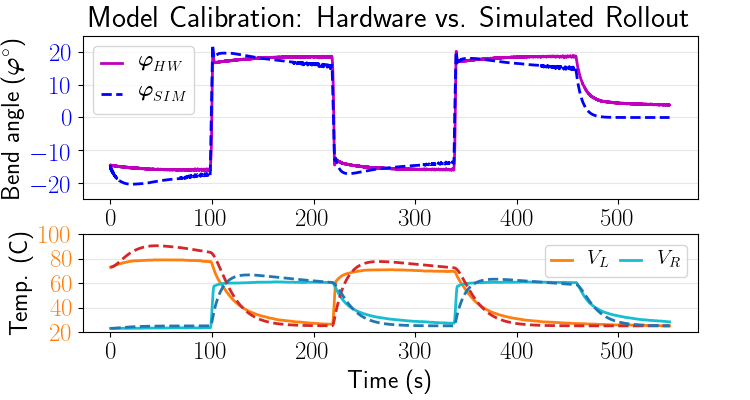}
    \caption{A simulation of the calibrated model (dashed) against a hardware test for the same inputs (solid) aligns qualitatively.}
    \label{fig:final_cal}
\end{figure}

With all parameters identified, the dynamics can be compared to the hardware dataset $\mathcal{C}^m$ that includes activation of both actuators.  
We simulated our model in open loop using the inputs $\vect{D}_{1 \hdots t}$ in $\mathcal{C}^m$, one subset plotted in Fig. \ref{fig:final_cal}.
Though many approximations were made in both the dynamics derivation and calibration, the simulation faithfully predicts the limb's motion.

%%%%%%%%%%%%%%%%%%%%%%%%%%%%%%%%%%%%%%%%%%%%%%%%%%%
%%%%%%%%%%%%%%%%%%%%%%%%%%%%%%%%%%%%%%%%%%%%%%%%%%%
%%%%%%%%%%%%%%%%%%%%%%%%%%%%%%%%%%%%%%%%%%%%%%%%%%%
\section{Trajectory optimization}

Using this calibrated model, we propose the following optimization routine that finds feasible state-input trajectories ($\vect{x}^*_{1\hdots N},\vect{u}^*_{1\hdots N}$) given an initial (likely infeasible) reference ($\vect{x}_{ref}$).
We define the state vector at time \(k\), \(\vect{x_k}\), with the joint angles \(\vect{\theta}\), measured wire temperatures \(V_l\) and \(V_r\), and their derivatives.
The input vector \(\vect{u_k} = \vect{D}\) contains PWM duty cycles for the left and right actuators:

\begin{equation}
    \vect{x} = \begin{bmatrix}
        \vect{\theta} & V_l & V_r & \vect{\dot\theta} & \dot V_l & \dot V_r 
    \end{bmatrix}^\top,
    \quad
    \vect{u} = \begin{bmatrix}
        D_l & D_r
    \end{bmatrix}^\top
    \label{eq:opt:state}
\end{equation}

Our nonlinear optimization program takes a direct collocation approach \cite{kelly2017introduction} to find $\vect{x}^*_{1\hdots N},\vect{u}^*_{1\hdots N}$, using a quadratic-cost objective (eqn. \ref{eq:opt}) and the constraints in eqns. (\ref{eq:opt:cinit})-(\ref{eq:opt:warm}).
We used IPOPT to solve the combined problem below, where \(N\) is the number of knot points, \(\vect{\tilde x_k}=\vect{x_k} - \vect{x_{k,ref}}\) is the deviation from the reference state at discrete time \(k\), and \(\vect{f}\) is the discrete dynamics function aggregated from the actuator and limb dynamics (Fig. \ref{fig:model:block}).
Objective weights were $\vect{Q} = 100 \; \text{diag}[\onevec{n}, \vect{0}]$ for the $n$-segment limb, only weighting $\vect{\theta}$, and $\vect{R} = 2 \vect{I}$. 
We do not include a terminal constraint to avoid an infeasible problem; instead, we use a large terminal weight $\vect{Q_N} = 1000 \vect{Q}$.

\begin{align}
    \vect{x}^*_{1\hdots N}, \vect{u}^*_{1 \hdots N} &= \arg\min_{\vect{x},\vect{u}} \frac{1}{2}\sum_{k=1}^{N-1}\left( \vect{\tilde x_k}^\top \vect{Q} \vect{\tilde x_k} + \vect{u_k}^\top \vect{R} \vect{u_k} \right) \notag \\
    & \qquad\qquad\, + \frac{1}{2}\vect{\tilde x_N}^\top \vect{Q_N} \vect{\tilde x_N} \label{eq:opt} \\
    \text{s.t.} & \quad \vect{x_1} = \vect{x_\text{init}} \label{eq:opt:cinit}\\
    & \quad \vect{x_{k+1}} = \vect{f}(\vect{x_k},\vect{u_k}) \label{eq:opt:cdyn}\\
    & \quad \vect{0} \le \vect{u_k} \le \onevec{2} \label{eq:opt:cctrl}\\
    & \quad \vect{T_k} < T_{max}\onevec{2} \label{eq:opt:safetyl} \\
    & \quad \vect{T_k} > T_{warm}\onevec{2} \qquad \forall k > k_{warm}\,. \label{eq:opt:warm}
\end{align}

Our constraints include the physical limits of PWM duty cycle (eqn. \ref{eq:opt:cctrl}) and a maximum of temperature (eqn. \ref{eq:opt:safetyl}), chosen as a conservative \(T_{max}=\SI{100}{\degreeCelsius}\) to prevent damage to our $90^\circ$C SMAs.
We also include a warmup constraint (eqn. \ref{eq:opt:warm}) with \(T_{warm}=\SI{45}{\degreeCelsius}\), \(k_{warm}= 20\)sec., since we observed a better model fit to actuator force at higher temperatures.

Note that the dynamics and optimization are expressed in joint space, not task space.
To determine the corresponding bend angle trajectory \(\phi^*_{1 \hdots N}\), the forward kinematics are used to compute the tip position \(\vect{t}_{1 \hdots N}\), and then \(\phi = \arctan(\vect{t}\cdot\vect{E_2}/\vect{t}\cdot\vect{E_1})\).
Lastly, the wire temperatures $\vect{T}$ for eqns. (\ref{eq:opt:safetyl})-(\ref{eq:opt:warm}) were calculated from the states $V_j$ in $\vect{x}$ via rearranging eqn. (\ref{eq:Vdot}) into $T_j = V_j - \dot V_j/a_3$.

%%%%%%%%%%%%%%%%%%%%%%%%%%%%%%%%%%%%%%%%%%%%%%%%%%%
%%%%%%%%%%%%%%%%%%%%%%%%%%%%%%%%%%%%%%%%%%%%%%%%%%%
%%%%%%%%%%%%%%%%%%%%%%%%%%%%%%%%%%%%%%%%%%%%%%%%%%%
\section{Results}

\begin{figure}[th] 
    \centering
    \includegraphics[width=1\columnwidth]{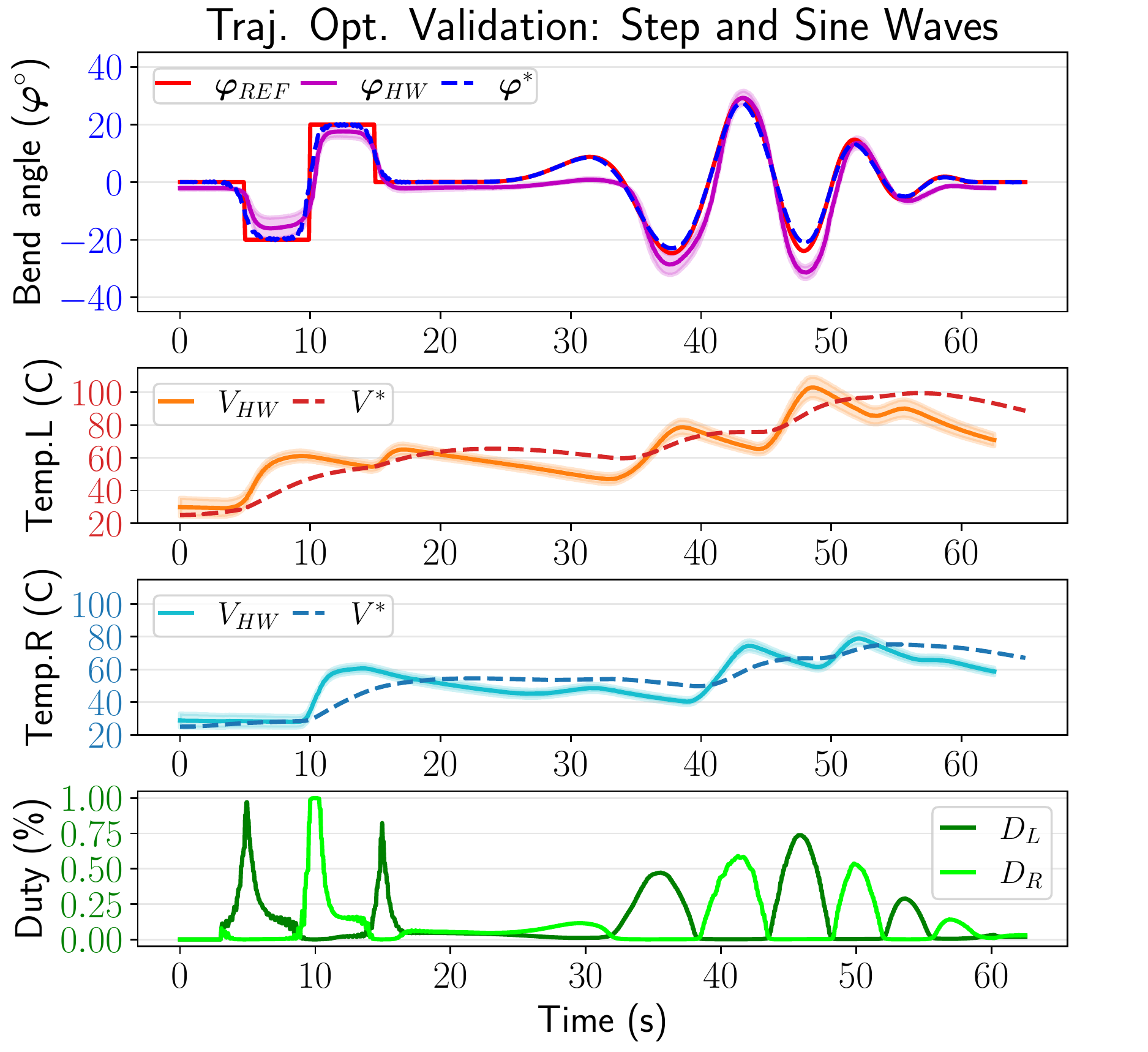}
    \caption{Trajectory optimization and hardware validation for step inputs and a decaying sine wave. Optimization produced the feasible state trajectory ${\phi^*}$ (blue dashed) including expected SMA measurement temperatures $V^*$ (dashed).
    Five hardware rollouts (\(\phi_{HW}\), magenta) qualitatively align with the optimized trajectory.}
    \label{fig:hw_val}
\end{figure}

We performed three experiments where a trajectory was optimized in software then executed in hardware.
Our first test serves as a validation of the concept, where we specified the $\phi_{REF}$ in Fig. \ref{fig:hw_val} by hand, consisting of two step inputs and a decaying sinusoid.
Converting $\phi_{REF} \rightarrow \vect{\theta}_{ref}$ via eqn. (\ref{eqn:beamcc}), we solved (\ref{eq:opt})-(\ref{eq:opt:warm}) to obtain the feasible trajectory $\phi^*$.
The corresponding open-loop inputs $\vect{u}^*$ were executed in hardware five times, plotted in Fig. \ref{fig:hw_val} as the mean result $(\cdot)_{HW}$ and a shaded 95\% confidence interval.

\begin{figure*}[thp]
    \centering
    \begin{subfigure}{1\columnwidth}
        \includegraphics[width=1\columnwidth]{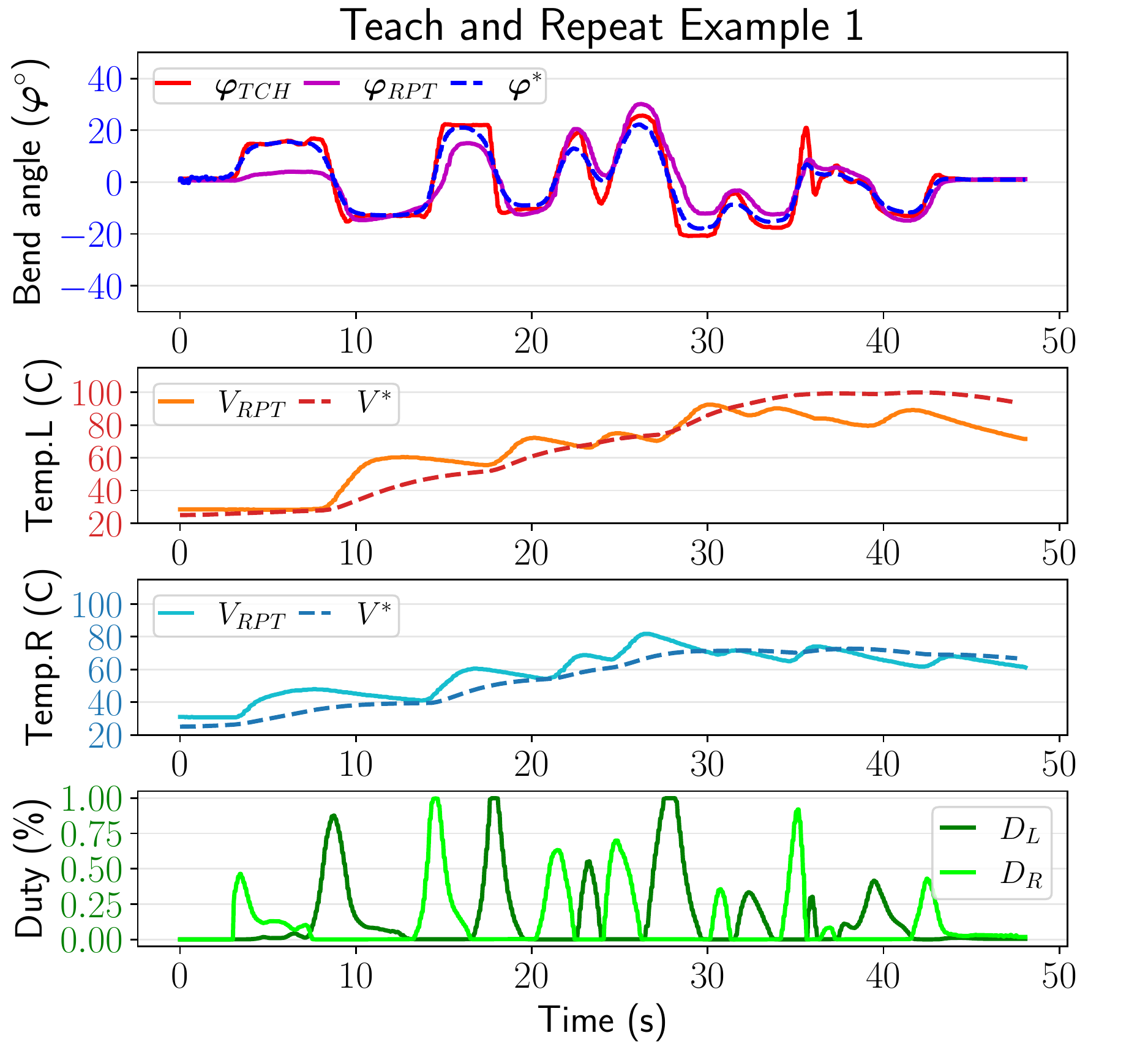}
    \end{subfigure}\quad
    \begin{subfigure}{1\columnwidth}
        \includegraphics[width=1\columnwidth]{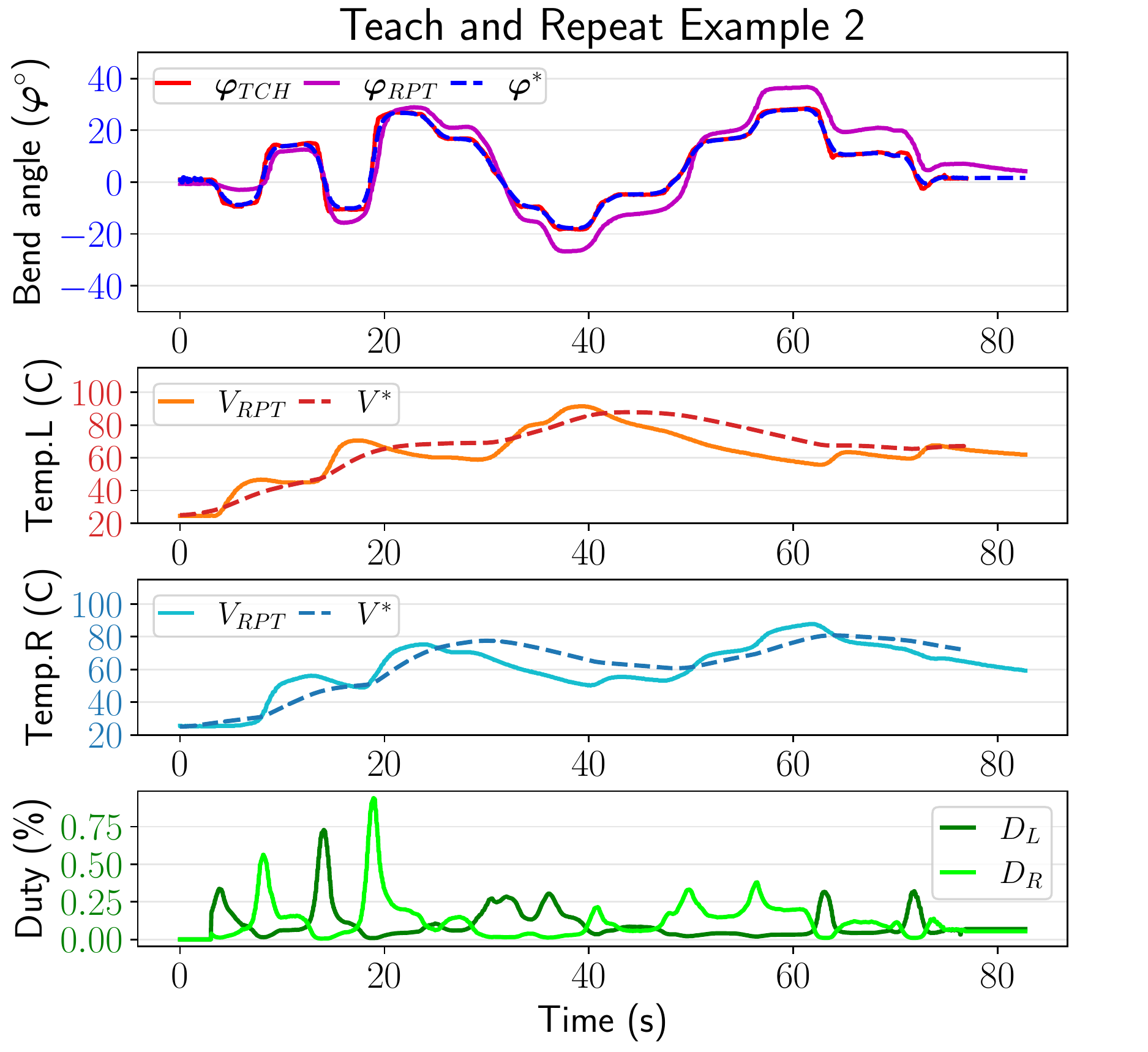}
    \end{subfigure}
    \caption{Two teach-and-repeat tests show that our method can recreate desired motions with relatively small error (Table \ref{tab:errors}). Temperatures (orange, cyan) are more dynamic in hardware, since we model $V$ with a lag.}
    \label{fig:teachrepeatresults}
\end{figure*}

\begin{figure}
    \centering
    \includegraphics[width=1\columnwidth]{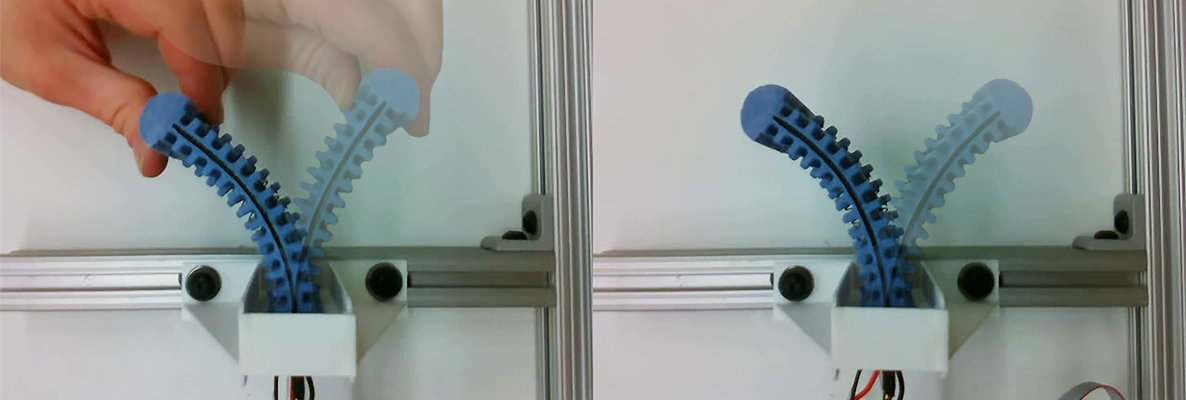}
    \caption{Teach and repeat tests show the use of our trajectory optimization procedure. Given an infeasible demonstration from moving the limb by hand (left), our optimization returns a feasible state/input result, which was tested open-loop in hardware (right).}
    \label{fig:teachrepeat_hw}
\end{figure}

The second two experiments were `teach and repeat' (T\&R) tests, where an initial trajectory $\phi_{TCH}$ was obtained by moving the limb by hand, with no actuation, and recording bend angle measurements (Fig. \ref{fig:teachrepeat_hw}).
These two dynamically infeasible $\phi_{TCH}$ were optimized to $\phi^*$ in Fig. \ref{fig:teachrepeatresults}, and as with the step/sinusoid test, the inputs $\vect{u}^*$ were executed in hardware. 
The first trajectory (`Example 1') tests faster motions within a small range of angles, whereas the second (`Example 2') tests a wider range of angles with small holds throughout.

All three experiments show that our procedure can recreate a variety of intricate motions.
The mean tracking error between $\phi^*$ and $\phi_{HW/RPT}$ remained relatively small (\ang{3} to \ang{5}, Table \ref{tab:errors}) in comparison to our inexpensive sensor's capabilities.
\textit{In situ} we typically observed around \ang{2} measurement error using this Bendlabs sensor, consistent with the accuracy noted in prior work~\cite{li_integration_2020}.
Tracking errors were larger in all tests in regions where either (a) the actuators were at low temperatures, during the warm-up period, (b) the desired angles were large, at the limits of our calibration range, or (c) the motions are very dynamic, and our calibration assumptions are violated.
Both overshoot (at fast motions or large angles) and undershoot (around $\phi=0$) were observed.

\begin{table}
    \caption{Tracking errors for the three hardware tests, expressed in absolute error $|\phi^* - \phi_{HW}|$.}
    \begin{center}
    \begin{tabular}{cccc}
        \hline
        \textbf{Trajectory} & \textbf{Mean} & \textbf{Median} & \textbf{90\% Percentile} \\
        \hline
        Validation & \ang{3.44} & \ang{2.33} &  \ang{7.43} \\
        T\&R 1     & \ang{3.90} & \ang{2.73} & \ang{10.61} \\
        T\&R 2     & \ang{5.27} & \ang{5.09} &  \ang{9.20} \\
        \hline
    \end{tabular}
    \label{tab:errors}
    \end{center}
\end{table}

The measured temperatures $V_{HW}$ were similar in magnitude to the optimized state trajectory $V^*$, but considerably more dynamic, indicating some mismatch in the relationship between the measurement (\(V\)) versus wire (\(T\)) temperatures.
This may be expected, since the calibration from Sec. \ref{sec:thermal_calibration} optimizes for bend angle, not temperature alignment.

%%%%%%%%%%%%%%%%%%%%%%%%%%%%%%%%%%%%%%%%%%%%%%%%%%%
%%%%%%%%%%%%%%%%%%%%%%%%%%%%%%%%%%%%%%%%%%%%%%%%%%%
%%%%%%%%%%%%%%%%%%%%%%%%%%%%%%%%%%%%%%%%%%%%%%%%%%%
\section{Discussion \& Conclusion}

This article demonstrates the first example of generating and optimizing motion trajectories for a soft thermally-actuated robot limb.
Our approach does not require external sensing nor computationally-challenging models, and open-loop control is simple to integrate with minimal electronics.
With this method, a single-segment limb performed open-loop tracking of a trajectory, including re-creation of `teach and repeat' motions.
In addition to showing proof-of-concept, the errors observed in Table \ref{tab:errors} are reasonably small in comparison to range of motion.
For example, with T\&R 2, median absolute error vs. angle range is $\approx \ang{5} / \ang{50} = 10\%$.
Since there is growing evidence that soft robots use \textit{embodied intelligence} to compensate for imprecise motions \cite{laschi_soft_2016}, our approach may be sufficient for many tasks such as locomotion \cite{patterson_untethered_2020}.

%%%%%%%%%%%%%%%%%%%%%%%%%%%%%%%%%%%%%%%%%%%%%%%%%%%
\subsection{Limitations}

The methodology in this article is designed for planar soft robots, and is only verified for a single-segment soft robot with two actuators.
A 3D implementation would require changing the calibration procedure, and multi-segment limbs pose challenges including greater computational complexity (with more discretized links) and propagation of modeling errors.
However, research exists on both 3D calibration \cite{hyatt_configuration_2019} and multi-segment soft robot modeling using a discretized manipulator \cite{graule_somo_2021}, which we may adapt for future work.

Our results show that average wire temperatures increase throughout the execution of a trajectory, mostly due to slow convective cooling of the thermal actuators.
Our approach may therefore be limited for long-term operation due to the $T_{max}$ constraint in eqn. (\ref{eq:opt:safetyl}).
Stiffness is also increasing, which may require increased power consumption, and sacrifices compliance.
Active cooling may reduce these effects~\cite{cheng2017modeling}, though at significant design cost, similar to issues with pneumatics or cable actuation.
Despite these limitations, our approach applies as-is to untethered soft robots with minor sensing additions (such as in \cite{patterson_robust_2021}) which also use SMAs.

%%%%%%%%%%%%%%%%%%%%%%%%%%%%%%%%%%%%%%%%%%%%%%%%%%%
\subsection{Sources of Error}

Our approximations, which make proof-of-concept trajectory optimization possible, also introduce modeling error.
The linear thermal actuator model is particularly simplistic, since the motion induced by a temperature change was nonlinear at low temperatures. 
The warm-up constraint in eqn. (\ref{eq:opt:warm}) only partially avoids this problem.
Using instead a constitutive model of shape-memory materials would capture the energy absorption due to phase change.
In addition, while the measured ($V)$ vs. wire ($T$) temperature model in eqns. (\ref{eq:Tdot})-(\ref{eq:Vdot}) was needed for our sensor design, it effectively adds a low-pass filter to temperature, causing less dynamic temperature predictions in the hardware tests (Fig. \ref{fig:hw_val}, \ref{fig:teachrepeatresults}).
However, our goal is task space ($\phi$) tracking performance, which showed similar response times between hardware and simulation.

Our test setup also inherently introduces imprecision.
The gravitational loading for the spring constant $k$ calibration routinely showed $\approx \ang{2}$ differences between hardware trials, vs. a \ang{10} total deflection.
Future work will improve calibration procedures.
Other assumptions, such as a known ambient temperature $T_0$, may be eliminated with more sensing.

%%%%%%%%%%%%%%%%%%%%%%%%%%%%%%%%%%%%%%%%%%%%%%%%%%%
\subsection{Future Work}

Ongoing work seeks to extend our approach in four major directions.
First, we seek to develop our method for 3D, multi-limbed robots.
Second, if and when this capability is developed, future work will use the approach to generate walking locomotion of e.g. the robot in \cite{patterson_untethered_2020}.
Third, we will examine closed-loop control strategies that may address model mismatch and unmodeled environments.
Techniques may include the time-varying linear quadratic regulator, iterative learning control, or even model-predictive control due to our model's computational simplicity.
Fourth, we plan to improve both the shape-memory model and hardware calibration routine as discussed above.
Combined, this approach shows potential both for use as-is in open-loop, as well as future applications with feedback.

Lastly, we gratefully acknowledge Zach J. Patterson for guidance in deriving eqn. (\ref{eqn:beamexponential}).

\bibliographystyle{IEEEtran_nourl}
\bibliography{IEEEabrv,sources.bib}

\end{document}